\documentclass[runningheads]{llncs}
\usepackage{graphicx}
\usepackage{mathtools}
\usepackage{multirow}
\usepackage[hyphens]{url}
\usepackage[table]{xcolor}
\usepackage{xcolor}

\begin{document}
\title{SplitFed resilience to packet loss: Where to split, that is the question\thanks{Supported in part by the Natural Sciences and Engineering Research Council (NSERC) of Canada under the grants RGPIN-2021-02485 and RGPAS-2021-00038.}}
%
%
\author{Chamani Shiranthika
\and
Zahra Hafezi Kafshgari
\and
Parvaneh Saeedi
\and
 Ivan V. Baji\'{c}
 }
\authorrunning{C. Shiranthika et al.}
%
\institute{School of Engineering Science, Simon Fraser University, Burnaby, BC, Canada 
\email{\{csj5, zha72, psaeedi, ibajic\}@sfu.ca}}
\maketitle              
\begin{abstract}
Decentralized machine learning has broadened its scope recently with the invention of Federated Learning (FL), Split Learning (SL), and their hybrids like Split Federated Learning (SplitFed or SFL). The goal of SFL is to reduce the computational power required by each client in FL and parallelize SL while maintaining privacy. This paper investigates the robustness of SFL against packet loss on communication links. The performance of various SFL aggregation strategies is examined by splitting the model at two points -- shallow split and deep split -- and testing whether the split point makes a statistically significant difference to the accuracy of the final model. Experiments are carried out on a segmentation model for human embryo images and indicate the statistically significant advantage of a deeper split point. 

\end{abstract}
\section{Introduction}
\label{sec:intro}
\vspace{-6pt}
Federated learning (FL)~\cite{mcmahan_2017} enables the training of machine learning models by multiple clients without sharing data. FL holds great promise for healthcare because of privacy constraints regarding medical data. In FL, clients train their local models and send them to the server for aggregation, after which the aggregated global model is sent back to the clients. Although FL addresses privacy concerns, it requires all clients to train local models that are usually of the same size as the global model. Since clients might not have the necessary computing resources (comparable to the server), this presents a challenge, especially for training large models.

Split Learning (SL)~\cite{Gupta_2018,Vepakomma_2018} was developed to overcome this client-server processing disparity.  In SL, a model is split into several parts that can reside in various locations and/or devices. Typically, the front-end of the model (usually the first few layers) is located on a client device, and the more computationally demanding back-end is located on a server.  During model training, features are sent from the front-end to the back-end, while gradients are sent from the back-end to the front-end. Thus, SL can solve the existing computational imbalance between the client(s) and the server in FL. However, SL on its own does not enable clients to collaborate in model training. Hence, recent research has blended FL and SL, resulting in hybrid Split-Federated Learning (SFL)~\cite{Thapa_2022,shiranthika_2023}, which combines the best of both worlds. SFL allows privacy preservation and collaboration between clients (like FL) while balancing computational resources between the client(s) and the server (like SL).

Error resilience is a critical challenge in distributed learning. The robustness of SFL to annotation errors has recently been studied in~\cite{Zahra_isbi_2023}, while the issue of noisy communication links was tackled in~\cite{Zahra_icassp_2023}. Packet loss is another frequent transmission error in real-world communication networks, which occurs when one or more data packets fail to reach their destination. Several attempts have been made to address packet loss in the FL literature.
Authors in~\cite{shirvani_2022} modeled the link between the clients and the server in FL as a packet erasure channel and experimentally studied the model convergence with and without packet loss. Loss tolerant FL (LT-FL) was explored in~\cite{zhou2021loss} in terms of aggregation, fairness, and personalization. Authors used ThrowRightAway (TRA) to accelerate the data uploading for low bandwidth devices by intentionally ignoring some packet losses. In SL, packet loss happens at model split points. Therefore, the question of \emph{where to split} directly impacts the loss resilience. The optimal choice of split points~\cite{kim_splitnet,tuli_splitplace_2022} and loss resilience~\cite{ALTeC_Access2020,Inpainting_ICC2021,CALTeC_ICIP2021} have been active, but thus far separate, research topics in \emph{split inference} (SI) or \emph{colloborative intelligence} (CI)~\cite{kang_neurosurgeon_2017,bajic2021collaborative}. However, to the best of our knowledge, there appear to be no existing studies of the impact of the choice of split points on loss resilience in SL, let alone the more recent SFL paradigm. 

This study investigates the impact of model split points on the loss resilience of SFL. We examine five parameter aggregation algorithms under various conditions such as different numbers of clients facing packet loss and different loss rates. Section~\ref{sec:system_model} describes the system model and the aggregation methods examined. Section~\ref{sec:experiments} describes the experiments and provides an analysis of the results. Conclusions and suggestions for future work are given in Section~\ref{sec:conclusions}.

\vspace{-10pt}
\section{System model}
\label{sec:system_model}
\begin{figure}[t]
\centerline{\includegraphics[scale = 0.4]{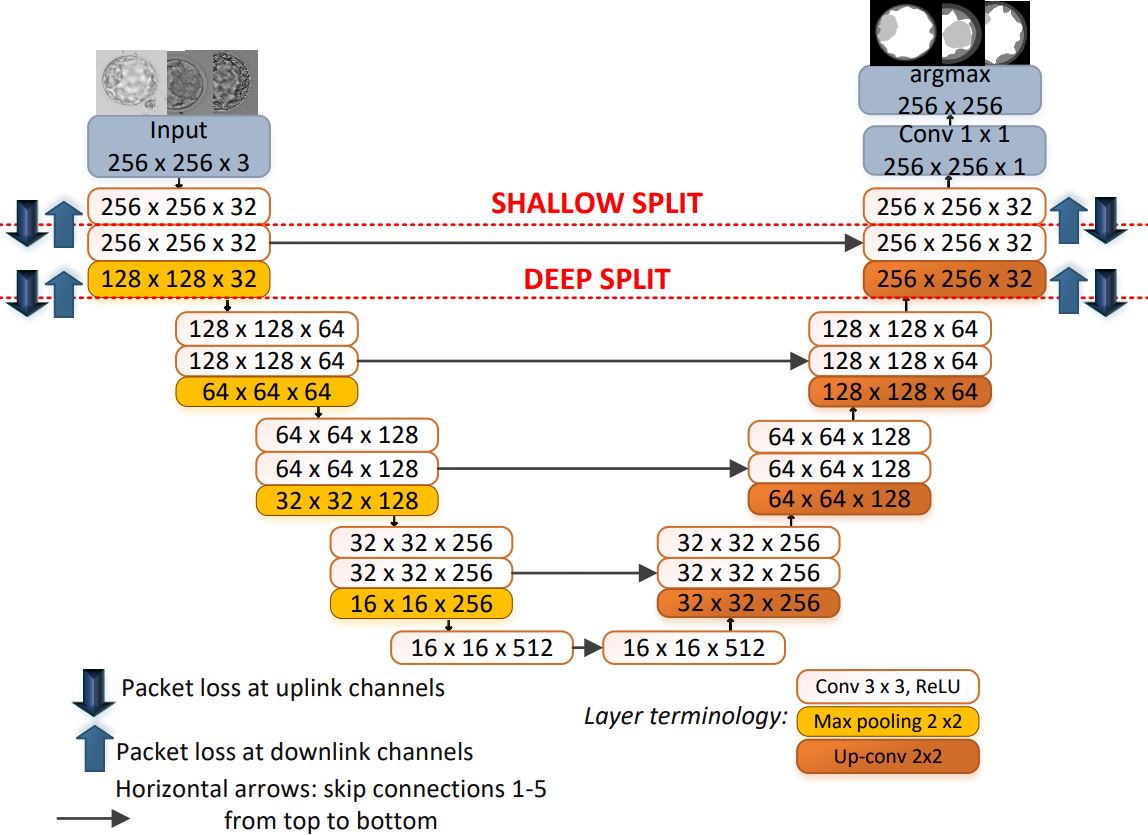}}
\caption{{\small{Split U-Net architecture}}}
\label{Splitted}
\end{figure}

\vspace{-6pt}
Fig.~\ref{Splitted} shows a SplitFed U-Net model for human embryo component segmentation on which our experiments are conducted. The U-Net consists of four downsampling blocks between the input and the bottleneck, and four upsampling blocks between the bottleneck and the output. Each block contains two convolutional layers with $3\times3$ kernels, a batch normalization layer, and \texttt{ReLU} activation. Each downsampling block starts with the aforementioned two convolutional layers followed by a $2\times2$ max-pooling layer. The number of filters in the four downsampling blocks increases as 32, 64, 128, and 256, from input towards the bottleneck. The bottleneck consists of two convolutional layers with 512 filters. Each upsampling block starts with a $2\times2$ upsampling layer followed by a transpose convolutional layer. The number of filters in the four upsampling blocks increases to 256, 128, 64, and 32 toward the output.  The final upsampling block is followed by a convolutional layer with the \texttt{argmax} function. 

We examine two ways of splitting the model: \emph{shallow split} and \emph{deep split}, indicated in Fig.~\ref{Splitted}. In shallow split, the first convolutional layer (front-end, FE), and the last two convolutional layers together with the output argmax layer (back-end, BE) are located on the client side, while the rest of the model is on the server. In deep split, the first two convolutional layers and the first max-pooling layer (front-end, FE), and the last three convolutional layers together with the final upsampling layer (back-end, BE) are located on the client side, while the rest of the model resides on the server.

\begin{figure}[t]
\centerline{\includegraphics[scale = 0.6]{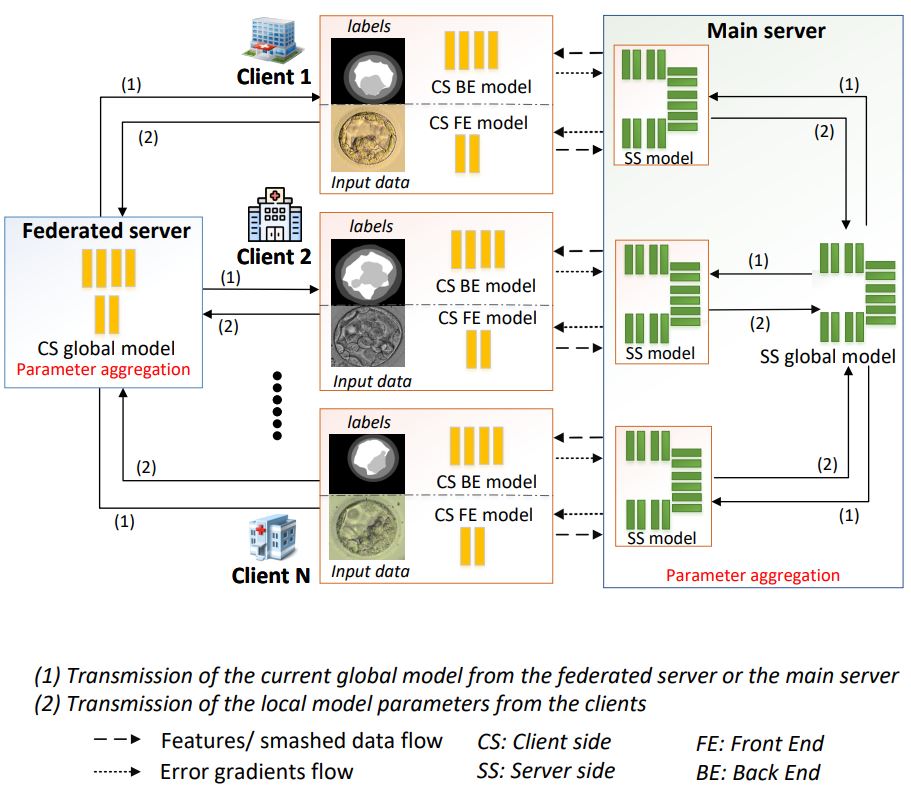}}
\caption{{\small{SplitFed U-Net architecture~\cite{shiranthika_2023} }}}
\label{Splitfed}
\end{figure}

The training process is as follows. First, initial copies of FE and BE are sent to each client, and the server makes a separate copy of its own model for each client. Each client then trains its local FE and BE in collaboration with its own copy of the server model for a certain number of local epochs. After that, each client sends its FE and BE models to the server, and aggregation is applied to all clients' FEs, BEs, and copies of the server model. The new aggregated global model consists of FE, server model, and BE. The server sends global FE and BE to each client to perform local validation. This completes one global epoch. During the forward pass, the features produced by the FE are sent from the client to the server. The server processes them through its own model and sends the resulting features back to the client to be processed by the BE. Client-side BE produces the prediction, computes the loss, and starts the back-propagation. Gradient updates from the client-side BE are sent to the server, back-propagated through the server model, and then sent to the client-side FE. 
Fig.~\ref{Splitfed} shows the adopted splitFed architecture of the U-Net model. 

We implemented five well-known parameter aggregation algorithms: naïve averaging~\cite{mcmahan_2017}, federated averaging (FedAvg)~\cite{mcmahan_2017}, auto-FedAvg~\cite{Xia_2021}, fed-NCL\_V2~\cite{Li_2021}, and fed-NCL\_V4~\cite{Li_2021}. In naïve averaging, parameter aggregation is based on the number of clients, while FedAvg considers the client's data distribution. Auto-FedAvg considers client's current training progress with their data distribution. In Fed-NCL\_V2, each layer gets the same weight, while in Fed-NCL\_V4, layer weights are proportional to their divergence from the global model. In both cases, parameter aggregation is based on the client's data distribution and local model divergence from the aggregated global model, while fed-NCL\_V2 additionally considers training loss of local training data on the aggregated global model.

\vspace{-8pt}
\section{Experiments}
\label{sec:experiments}
\vspace{-3pt}
\subsection{Experimental setup}
\begin{figure}[t]
\centerline{\includegraphics[scale=0.4]{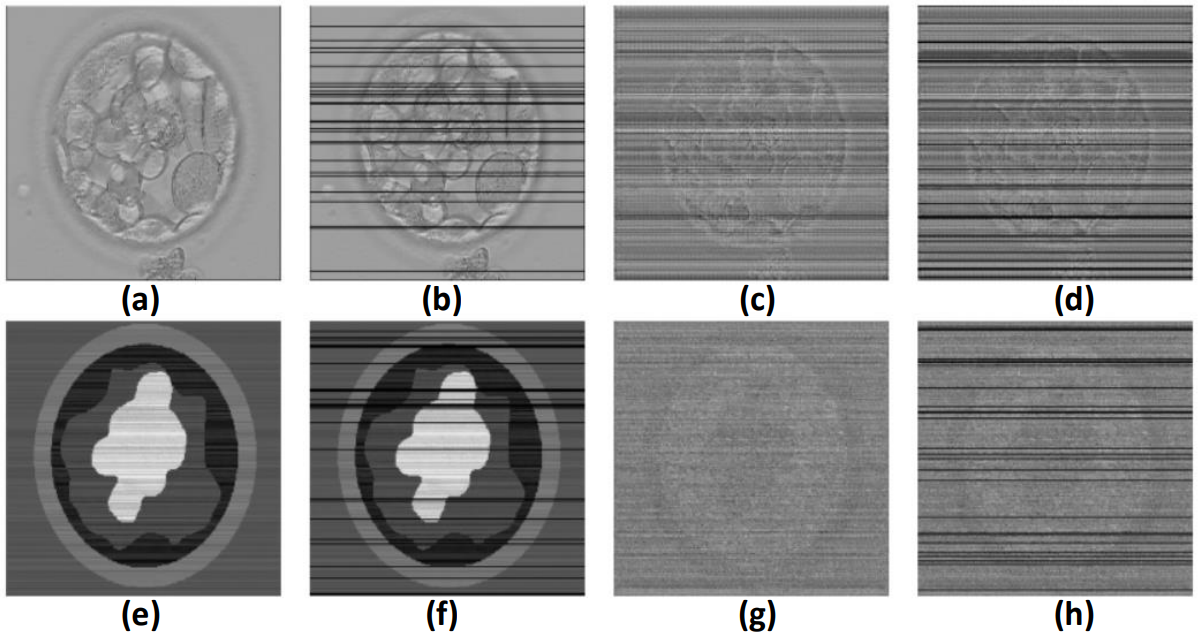}}
\caption{{\small {A feature map (top row) and gradient map (bottom row) subject to 10\% packet loss. Missing data is indicated by black horizontal lines. (a) Client FE feature map output before loss; (b)  Server input after loss; (c) Server output feature map; (d) Client BE input after loss; (e) Client BE output gradient map; (f) Server input after loss; (g) Server output gradient map; (h) Client FE input after loss. 
}}} 
\label{PL_ex1}
\end{figure}

The dataset consists of 781 human embryo images~\cite{lockhart_2019}, each with ground-truth segmentation masks for five components: Background, Zona Pellucida (ZP), Trophectoderm (TE), Inner Cell Mass (ICM), and Blastocoel (BL). Of these, 70 images are saved as the test set, and the rest are used for training. Data are distributed among 5 clients - 240, 120, 85, 179, 87. Each client reserves 85\% of its data for training and 15\% for validation. During training, the input images are resized to $256\times256$. Augmentation is performed using horizontal and vertical flipping. \emph{Soft Dice} loss~\cite{sudre_2017} is chosen as the loss function. Adam optimizer is used with the initial learning rate of ${10}^{-4}$. \emph{Mean Jaccard index} (MJI)~\cite{Cox_2008} without background is taken as the performance metric.
The system is trained for 12 local and 15 global epochs.
As in~\cite{ALTeC_Access2020}, each packet is assumed to be one row of a feature map. Data from the lost packets are assumed to be zero (i.e., no sophisticated packet loss concealment is deployed), so packet loss results in zeroing out some rows in the feature maps and gradient maps at split points. Fig.~\ref{PL_ex1} shows an example of a feature map in the forward pass and a gradient map in the back-propagation pass in the first epoch of SFL, both
subject to 10\% packet loss. 

\vspace{-10pt}
\subsection{Baseline experiments without packet loss}
First, we verify the performance of our core U-Net model by comparing
it with BLAST-NET~\cite{Rad_2019}, a state-of-the-art network for human embryo image segmentation. We trained our U-Net model without splits, in a centralized manner\footnote{That is, without distributing data across the clients.} on the same dataset~\cite{Saeedi_2017} of 235 images that BLAST-NET was trained on. The MJI of our U-Net was 81.70\%, while the MJI of BLAST-NET~\cite{Rad_2019} is 79.88\%. Hence, our model compares favorably against BLAST-NET. Instead of achieving the new best embryo segmentation result, the aim of this experiment was simply to show that our U-Net model is a reasonable one.

Next, we verify the performance of our model in a split-federated scenario without packet loss. We test the performance of all five aggregation methods over 10 runs. Average MJI were 82.78\%, 82.57\%, 82.99\%, 83.02\%, and 82.95\% for naïve avg, FedAvg, auto-FedAvg, fed-NCL\_V2 and fed-NCL\_V4, respectively. 

We performed pairwise statistical significance testing for the difference in these average MJIs. Specifically, if $J_{\text{method1}}$ and $J_{\text{method2}}$ are MJI's of two parameter aggregation methods, the two-tail t-test is
\begin{equation}
    \begin{aligned}
        H_0: J_{\text{method1}} = J_{\text{method2}} \\
        H_1: J_{\text{method1}} \neq J_{\text{method2}}  
    \end{aligned}
\end{equation}
When the p-value~\cite{fisher_1936} is less than 0.05, the null hypothesis $H_0$ can be rejected (at the significance level of 0.05) to conclude that the difference is significant.

Most of the MJI differences were not statistically significant (p $\geq$ 0.05), except that FedAvg had a significantly lower MJI than auto-FedAvg, fed-NCL\_V2, and fed-NCL\_V4. This is not surprising, as all three methods were developed to improve over FedAvg.

\vspace{-10pt}
\subsection{Experiments with packet loss}

\begin{figure*}[t]
\centerline{\includegraphics[scale=0.56]{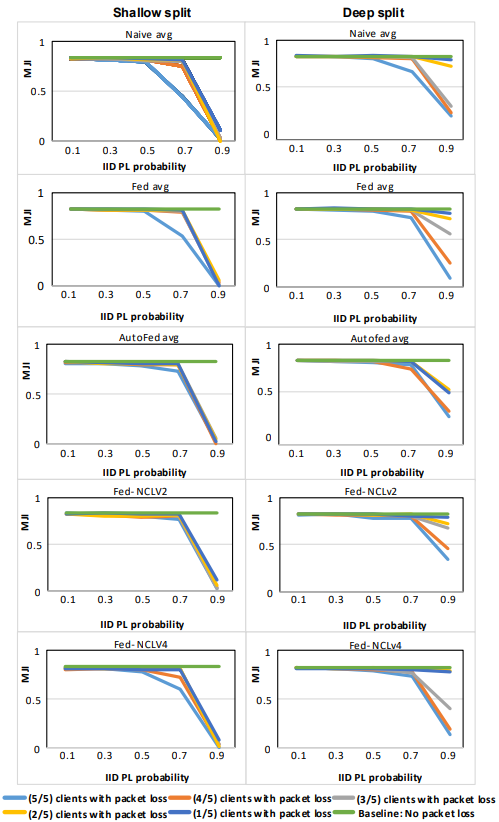}}
\caption{ {\small MJI vs. $P_L$ for shallow split (left) and deep split (right), with various numbers of clients experiencing packet loss.} }
\label{deepvsshallow_ex1}
\vspace{-10pt}
\end{figure*}

In our experiments, packet loss is assumed to be independent and identically distributed (iid) with probability $P_L\in\{0.1, 0.3, 0.5, 0.7, 0.9\}$. 
Fig.~\ref{deepvsshallow_ex1} shows the average MJI of the final trained model vs. $P_L$ over 10 runs for shallow- and deep-split models. In each case, six curves are shown: baseline performance without packet loss (green horizontal line) and the curves for $m/5$ clients experiencing packet loss, where $m\in\{1,2,...,5\}$. For $P_L\in\{0.1,0.3,0.5\}$, deep and shallow split curves are close to the performance without packet loss, regardless of how many clients are experiencing packet loss. When $P_L=0.7$, MJI starts to decrease, and more so when a larger number of clients experience packet loss.  When $P_L=0.9$, the shallow-split model ends up with near-zero MJI, regardless of how many clients experience packet loss. Meanwhile, the deep-split model can still be trained close to its no-loss performance when only a single client is experiencing packet loss, but in all cases ends up with higher MJI values than the shallow-split model.

Based on Fig.~\ref{deepvsshallow_ex1}, it appears that the deep-split model can be trained to a higher MJI than the shallow-split model under all conditions. To test this, we performed 125 pairwise t-tests comparing shallow vs. deep split, for each unique combination of $P_L$ and the number of clients experiencing packet loss. Table~\ref{tab:shallow_vs_deep_loss} shows the p-values of the corresponding one-tailed t-test comparing the MJI with deep and shallow splits,  $J_{\text{deep}}$ and $J_{\text{shallow}}$, respectively:
\begin{equation}
    \begin{aligned}
        H_0: J_{\text{deep}} \leq J_{\text{shallow}} \\
        H_1: J_{\text{deep}} > J_{\text{shallow}} \\
    \end{aligned}
\end{equation}
As seen in the table, in all cases we have p $<$ 0.05, so one can reject the null-hypothesis $H_0$ and conclude that deep split produces a higher MJI than shallow split at the significance level of 0.05. Moreover, table cells highlighted in green indicate the cases where p $<$ 0.01, and in all these cases, we can conclude that deep split is better than shallow split at a stronger significance level of 0.01. Hence, \emph{in SFL over lossy links, the split point has a significant influence on the final model performance, and a deeper split is better.} 

\begin{table}[t]
\centering
\small
\caption{{\small Summary of the pairwise t-tests \textbf{between shallow and deep split} under various conditions. Values less than 0.01 are highlighted in green.} 
}

\begin{tabular}{|c|c|c|c|c|c|c| }
\hline
\multicolumn{1}{|l|}{ \textbf{\shortstack{Parameter \\ aggregation \\ algorithm} }} & {\textbf{\shortstack{ \# \\clients \\  w/ loss}}} & \multicolumn{5}{c|}{\shortstack{\textbf{
One-tail p-value} \\ rounded off to $2^{\text{nd}}$ decimal place}}  \\ \hline
\multicolumn{2}{|c|} {$P_L$} & \quad0.1  \quad    &  \quad0.3  \quad    & \quad0.5  \quad    & \quad0.7 \quad    & \quad0.9  \quad    \\ \cline{1-7}

\multirow{5}{*}{\rotatebox[origin=c]{90}{Naïve avg}}  & 5 & 0.01 &	0.02	&0.02&	0.02&	0.02 \\\cline{2-7} 
&  4  & \cellcolor{green!15}0.00	& \cellcolor{green!15}0.00 &	\cellcolor{green!15}0.00 &	\cellcolor{green!15}0.00 &	\cellcolor{green!15}0.00\\ \cline{2-7}
&   3& \cellcolor{green!15}0.00 &	0.03 &	\cellcolor{green!15}0.00 &	0.02 & \cellcolor{green!15}0.00 \\ \cline{2-7} 
&   2  & \cellcolor{green!15}0.00 &	\cellcolor{green!15}0.00 &	0.02 &	0.02 &	\cellcolor{green!15}0.00\\ \cline{2-7} 
&   1   &\cellcolor{green!15}0.00 & \cellcolor{green!15}0.00 &	0.02 &	0.03 &	\cellcolor{green!15}0.00\\ \cline{2-7} \hline
  
\multirow{5}{*}{\rotatebox[origin=c]{90}{Fed avg}} & 5 & \cellcolor{green!15}0.00 & \cellcolor{green!15}0.00 &	\cellcolor{green!15}0.00 &	0.02 &	0.03\\\cline{2-7} 
&  4  & \cellcolor{green!15}0.00 &	\cellcolor{green!15}0.00 & \cellcolor{green!15}0.00 &	0.02 &	0.02\\ \cline{2-7} 
&   3& 0.01 &	0.03 &	\cellcolor{green!15}0.00 & \cellcolor{green!15}0.00 &	\cellcolor{green!15}0.00 \\ \cline{2-7} 
&   2  & \cellcolor{green!15}0.00 &	\cellcolor{green!15}0.00 &	0.01 &	0.03 &	\cellcolor{green!15}0.00 \\ \cline{2-7} 
&   1   &\cellcolor{green!15}0.00 & \cellcolor{green!15}0.00 &	\cellcolor{green!15}0.00 &	\cellcolor{green!15}0.00 &	\cellcolor{green!15}0.00\\ \cline{2-7} \hline
                    
\multirow{5}{*}{\rotatebox[origin=c]{90}{Auto-Fedavg}} & 5 & 0.02 &	0.01 &	\cellcolor{green!15}0.00 &	\cellcolor{green!15}0.00 &	\cellcolor{green!15}0.00 \\\cline{2-7} 
&  4  & \cellcolor{green!15}0.00 &	\cellcolor{green!15}0.00 	&\cellcolor{green!15}0.00 &	\cellcolor{green!15}0.00 &	\cellcolor{green!15}0.00\\ \cline{2-7} 
&   3& \cellcolor{green!15}0.00 &	\cellcolor{green!15}0.00 &	\cellcolor{green!15}0.00 &	\cellcolor{green!15}0.00 &\cellcolor{green!15}0.00 \\ \cline{2-7} 
&   2  & 0.02 &	\cellcolor{green!15}0.00 &	\cellcolor{green!15}0.00 & \cellcolor{green!15}0.00&	\cellcolor{green!15}0.00 \\ \cline{2-7} 
&   1   &\cellcolor{green!15}0.00 & \cellcolor{green!15}0.00 & \cellcolor{green!15}0.00 &	0.03 &	\cellcolor{green!15}0.00\\ \cline{2-7} \hline
                   
\multirow{5}{*}{\rotatebox[origin=c]{90}{Fed-NCL\_V2}} &  5 & \cellcolor{green!15}0.00 &	\cellcolor{green!15}0.00 &	0.01 &	\cellcolor{green!15}0.00 &	\cellcolor{green!15}0.00\\\cline{2-7} 
&  4  & 0.01 &	0.02 &	0.02 &	\cellcolor{green!15}0.00 &	\cellcolor{green!15}0.00\\ \cline{2-7} 
&   3& \cellcolor{green!15}0.00 &	\cellcolor{green!15}0.00 &	\cellcolor{green!15}0.00 &	\cellcolor{green!15}0.00 &	\cellcolor{green!15}0.00 \\ \cline{2-7} 
&   2  & \cellcolor{green!15}0.00 &	0.01 &	\cellcolor{green!15}0.00 &	0.01 &	\cellcolor{green!15}0.00 \\ \cline{2-7} 
&   1   &\cellcolor{green!15}0.00&	0.01 &	0.02 &	\cellcolor{green!15}0.00 &	\cellcolor{green!15}0.00\\ \cline{2-7} \hline
                   
\multirow{5}{*}{\rotatebox[origin=c]{90}{Fed-NCL\_V4}} & 5 & \cellcolor{green!15}0.00 &	0.02 &	\cellcolor{green!15}0.00&	\cellcolor{green!15}0.00 &	\cellcolor{green!15}0.00 \\\cline{2-7} 
&  4  & \cellcolor{green!15}0.00 &	\cellcolor{green!15}0.00 & \cellcolor{green!15}0.00 & \cellcolor{green!15}0.00 &\cellcolor{green!15}0.00\\ \cline{2-7} 
&   3& \cellcolor{green!15}0.00 &	\cellcolor{green!15}0.00 &	\cellcolor{green!15}0.00 &	\cellcolor{green!15}0.00 &	\cellcolor{green!15}0.00 \\ \cline{2-7} 
&   2  & \cellcolor{green!15}0.00 &	\cellcolor{green!15}0.00 &	\cellcolor{green!15}0.00 &	\cellcolor{green!15}0.00 & \cellcolor{green!15}0.00 \\ \cline{2-7} 
&   1   &\cellcolor{green!15}0.00 &	\cellcolor{green!15}0.00 &	\cellcolor{green!15}0.00 &	\cellcolor{green!15}0.00 &	\cellcolor{green!15}0.00\\ \cline{2-7} \hline
         
\end{tabular}
\label{tab:shallow_vs_deep_loss}
\end{table}
 
Finally, we examined whether any aggregation methods perform significantly better than others under the packet loss scenario for the deep-split model. With five aggregation methods, $\binom{5}{4} = 10$ pairwise comparisons can be made for five values of $P_L$ and five numbers of clients experiencing packet loss; hence $10\times5\times5 = 250$ comparisons. We performed a t-test for each of the $250$ cases. Some methods performed (significantly) better than others in certain cases, but we did not notice any pattern that would allow us to conclude that a certain method is better than others across the board. The full results can be found at \url{https://drive.google.com/drive/u/0/folders/140f6OGYLRhjqcNQe2aLbnfy1dA7dYt60.}

\vspace{-10pt}
\section{Conclusions and future work}
\label{sec:conclusions}
\vspace{-8pt}
In this paper, we examined the effects of model split points in split-federated learning (SFL) under packet loss. Experiments with five state-of-the-art aggregation methods showed that the split point has a statistically significant impact on the final model performance and that a deeper split is better. The reason for this is twofold: (1) the deep-split model has more layers available in the client-side back-end to learn how to recover the lost data, and (2) in our deep-split U-Net model, the first skip connection was fully located at the client and was able to transfer some features without packet loss.

It was also observed that SFL with our U-Net model was fairly robust to packet loss of up to 50\%, with both shallow and deep split. This can be due to two reasons. The first reason is the use of  \texttt{ReLU} activations in our split U-Net. It was reported in~\cite{ALTeC_Access2020} that models with \texttt{ReLU} activations tend to be fairly robust to packet loss, because \texttt{ReLU} activations produce a lot of zeros in their output. Hence, when a missing feature value is replaced with a zero, much of the time, the zero value is the actual value that was lost. 
On the other hand, many high-performance models for applications in medical image analysis and computer vision use other activation functions, and such models could be more sensitive to packet loss. 
The second reason is that packet loss can act as a regularization technique, similar to dropout. To test this, we compared the MJI of models trained under packet loss with $P_L\in\{0.1, 0.3, 0.5, 0.7, 0.9\}$ and models trained with dropout rates that match these values. A split U-Net model was trained on all training samples ten times for each $P_L$ and a matching dropout rate at the split points. At the significance level of 0.01, there were no statistically significant differences between the MJI of the models trained with packet loss and dropout of $P_L\in\{0.1, 0.3, 0.5\}$. For higher loss rates, models trained under packet loss had significantly lower MJI than those trained with the dropout. Hence, for low to moderate loss rates, the effect of packet loss is similar to dropout and will not negatively affect the MJI of trained models.

Other avenues for future work include testing the effectiveness of SFL with multiple application scenarios that apply diverse semantic segmentation networks across multiple split points, studying SFL with more realistic packet loss models, such as bursty loss or real packet traces, as well as developing more robust parameter aggregation algorithms for SFL and methods for packet loss recovery. Some work on missing data recovery in feature maps has been done in the context of collaborative inference~\cite{ALTeC_Access2020,Inpainting_ICC2021,CALTeC_ICIP2021}. However, in SFL, a loss is observed not only in feature maps but also in gradient maps, creating a new challenge. As the first study on the effects of packet loss in SFL, we hope that this paper will stimulate further work on that topic.

%
%
%
%

\end{document}